\documentclass[conference]{IEEEtran}

\usepackage{booktabs} 
\usepackage{amsmath}
\usepackage{adjustbox}
\usepackage{multirow}
\usepackage{url}
\usepackage{graphicx}
\usepackage{subfigure}
\usepackage{multirow}
\usepackage{subfigure}
\usepackage{bm}
\usepackage{balance}
\usepackage{amssymb}
\usepackage{pgfplots}
\usepackage{pifont}
\usepackage[usenames,dvipsnames]{pstricks}
\usepackage{epsfig}
\usepackage{booktabs}
\usepackage{pgffor}
\usepackage{epsfig,epstopdf}
\usepackage{stmaryrd}
\usepackage{gensymb}

\newcommand{\ie}[0]{\textit{i.e., }}
\usepackage{algorithm}
\usepackage{algorithmicx}
\usepackage{algpseudocode}
\usepackage{ mathrsfs }
\usepackage{comment}
\usepackage{pgfplots}
\pgfplotsset{compat=newest}
\IEEEoverridecommandlockouts    

\begin{document}

\title{Community-preserving Graph Convolutions for Structural and Functional Joint Embedding of Brain Networks}

\author{\IEEEauthorblockN{Jiahao Liu$^*$}
\thanks{$^*$Authors have equal contributions.$^{\dagger}$This work was done when the author was at the University of Illinois at Chicago.}
\IEEEauthorblockA{\textit{Department of Computer Science} \\
\textit{Tongji University}\\
Shanghai, China\\
jiahaoliu1891@gmail.com\\[0.1cm]
}
\IEEEauthorblockN{Chun-Ta Lu}
\IEEEauthorblockA{\textit{Google Research} \\
Mountain View, CA, USA\\
chunta@google.com}

\and
\IEEEauthorblockN{Guixiang Ma$^*$$^{\dagger}$}
\IEEEauthorblockA{\textit{Intel Labs} \\
Hillsboro, OR, USA\\
guixiang.ma@intel.com\\[0.5cm]
}
\IEEEauthorblockN{Philip S. Yu}
\IEEEauthorblockA{\textit{Department of Computer Science} \\
\textit{University of Illinois at Chicago}\\
Chicago, IL, USA\\
psyu@uic.edu
}
\and
\IEEEauthorblockN{Fei Jiang}
\IEEEauthorblockA{\textit{Computational and Applied Mathematics}\\
\textit{The University of Chicago} \\
Chicago, IL, USA\\
feijiang@uchicago.edu\\[0.1cm]
}
\IEEEauthorblockN{Ann B. Ragin}
\IEEEauthorblockA{\textit{Department of Radiology} \\
\textit{Northwestern University}\\
Chicago, IL, USA\\
ann-ragin@northwestern.edu
}
}


\maketitle

\begin{abstract}
Brain networks have received considerable attention given the critical significance for
understanding human brain organization, for investigating
neurological disorders and for clinical diagnostic applications. Structural brain network (e.g. DTI) and functional brain network (e.g. fMRI) are the primary networks of interest. Most existing works in brain network analysis focus on either structural or functional connectivity, which cannot leverage the complementary information from each other. Although multi-view learning methods have been proposed to learn from both networks (or views), these methods aim to reach a consensus among multiple views, and thus distinct intrinsic properties of each view may be ignored. How to jointly learn representations from structural and functional brain networks while preserving their inherent properties is a critical problem.


In this paper, we propose a framework of Siamese community-preserving graph convolutional network (SCP-GCN) to learn the structural and functional joint embedding of brain networks. Specifically, we use graph convolutions to learn the structural and functional joint embedding, where the graph structure is defined with structural connectivity and node features are from the functional connectivity. Moreover, we propose to preserve the community structure of brain networks in the graph convolutions by considering the intra-community and inter-community properties in the learning process. Furthermore, we use Siamese architecture which models the pair-wise similarity learning to guide the learning process.
To evaluate the proposed approach, we conduct extensive experiments on two real brain network datasets. The experimental results demonstrate the superior performance of the proposed approach in structural and functional joint embedding for neurological disorder analysis, indicating its promising value for clinical applications.

\end{abstract}


\section{Introduction} \label{sec_intro}
In recent years, advances in neuroimaging technology have given rise to various modalities of brain imaging data, which provide us with unprecedented opportunities for investigating the inner organization and activity of human brains. Techniques such as functional magnetic resonance imaging (fMRI), for example, can be used to study the functional activation patterns of human brain based on the cerebral blood flow and the BOLD response \cite{ogawa1990brain, cox2003functional}, while techniques like diffusion tensor imaging (DTI) can be used for examining the tractography of the white matter fiber pathways and thus for exploring the structural connectivity in the brain. 

In the past decade, brain networks derived from these brain imaging data have been widely studied for neurological disorder analysis, which is a critical problem in the field of biomedical neuroscience \cite{sporns2011human,kong2014brain}. Structural brain networks derived from DTI brain data and functional brain networks derived from fMRI brain data are two major kinds of brain networks that are often used in these studies. In fMRI brain networks, the edges encode the correlations between the functional activities of two different brain regions, while the links in DTI brain networks represents the number of neural fibers connecting the different regions. The connectivity structure in these brain networks can encode tremendous information about the neurological health status of the individuals, and thus it has been used widely in brain network analysis for neurological disorder diagnosis.  

Most existing works in brain network analysis focus on either the analysis of structural connectivity or on functional connectivity \cite{van2010exploring,beare2017altered,wang2017structural}. However, both the anatomical characteristics captured by structural connectivity and the physiological properties that form the basis of functional connectivity are important to understand the integrated organization and brain activity, and thus it would be of considerable benefit if both structural and functional networks could be considered in combination \cite{sporns2013structure}. Recently, some works have used multi-view embedding methods to learn representations from both structural and functional networks for neurological disorder analysis \cite{ma2017multiview,ma2017multi,zhang2018multimodal}. For example, \cite{ma2017multiview} proposes a multi-view graph embedding approach which learns a unified network embedding from functional and structural brain networks as well as hubs for brain disorder analysis.
\cite{zhang2018multimodal} proposes to fuse the multimodal brain networks along with the longitudinal information to learn a latent representation of the brain networks for predicting anxiety and depression. 
Although multi-view embedding methods consider both structural and functional networks, representations derived from each view in these methods are enforced to reach consensus, making distinct intrinsic properties of individual network being ignored. Therefore, it is desirable to find a way to jointly learn from structural and functional brain networks while considering their intrinsic properties. 

In this paper, we propose to use graph convolutional network (GCN) for learning the embedding of brain network by jointly using the structural brain network and the functional brain network. GCN, as a generalized convolutional neural network from grid-structure domain to graph-structure domain, has been emerging as a powerful approach for graph mining \cite{bruna2013spectral,defferrard2016convolutional,zhou2018graph,zhang2018deep}. Generally, GCN takes two inputs: a representative description of the graph structure in matrix form (e.g. adjacency matrix) and a feature vector for every node in the graph. The convolutions are then performed based on the neighborhood structure indicated by the given graph structure. 
Although GCN has been shown to be effective for network representation learning \cite{zhou2018graph}, there are some challenges that we need to address when using it to learn the structural and functional joint embedding of  brain network embedding for neurological disorder analysis, as listed below:

\begin{itemize}
\item \textbf{Joint embedding of DTI and fMRI}: The structural connectivity in DTI reflects the anatomical pathways of white matter tracts connecting different regions, whereas the functional connectivity in fMRI encodes the correlation between the activity of brain regions. Therefore, they cannot be fused directly or be treated in the same manner in GCN. How to apply GCN on the DTI brain network and the fMRI network jointly, so that the resulted embedding could encode the inherent properties of both structural connectivity and functional connectivity, is a key problem.
\item \textbf{Community-structure preserving}: Modular/community structure, as one of the key properties of brain networks, has been shown as an important factor in neurological disorder analysis \cite{sporns2013structure, rubinov2011weight, wang2017structural,ma2017multiview}. It is crucial to preserve the community structure while learning the embedding of brain network. However, existing GCN methods do not consider the community structure. How to make the graph convolutions be able to preserve the community structure for brain network analysis is a challenging problem. 
\item \textbf{Limited sample quantity}: Training a deep learning model requires a large amount of training data, but neurological disorder analysis often suffers from data scarcity problem, due to the high cost of the data acquisition and limited number of available patients of interest. How to address this problem is another key issue when using GCN for embedding learning. 
\item \textbf{Neurological Disorder Analysis}: How to leverage the brain network joint embedding learned by the community-preserving graph convolutions to facilitate neurological disorder analysis is also a critical problem.
\end{itemize}



To address these challenges, we propose a framework of Siamese community-preserving graph convolutional networks for learning the structural and functional joint embedding of brain networks. Our contributions can be summarized as follows: 
\begin{itemize}
     \item We propose to use graph convolutions for learning the joint embedding of fMRI functional brain network and DTI structural brain network, where DTI network defines the graph structure and fMRI network is used as node features in the convolutions. By considering the structural and functional networks jointly in this way, both the inherent structural information and the functional patterns can be captured and leveraged in the learning process.

     \item We propose to incorporate the community-preserving property into graph convolutional networks to preserve the intrinsic modular/community structure of human brain networks while learning their structural and functional joint embedding.
     
    \item We use the Siamese architecture \cite{koch2015siamese} and exploit pair-wise similarity learning of brain networks to guide the learning process, which could help alleviate the data scarcity problem.
    
    \item We integrate the community-preserving property and the pair-wise similarity learning strategy into a unified framework called "Siamese community-preserving GCN" (SCP-GCN) for structural and functional joint embedding of brain networks. Specifically, we propose a community-preserving loss and then incorporate it into the loss of the pair-wise learning in Siamese GCN to facilitate the community-preserving graph convolutions. Both the intra-community property and inter-community property are considered when formulating the community-preserving loss.
    
    \item We apply the proposed framework on two real brain network datasets (i.e., Bipolar and HIV \cite{ma2017multi}) to learn the structural and functional joint embedding for the detection of these two disorders. The experimental results demonstrate the superior performance of the proposed approach in structural and functional joint analysis for clinical investigation and application. 
\end{itemize}

The rest of this paper is organized as follows. The problem formulation and some preliminary knowledge are given in the next section. Then we present the proposed framework in Section \ref{sec_framework}. The experimental results and analysis are shown in Section \ref{sec_experiment}. Related works are discussed in Section \ref{sec_related} and the conclusions are given in Section \ref{sec_conclusion}.


\section{Preliminaries}
In this section, we first introduce the notations and terminologies we will use in this paper. Then we formulate our problem formally and present some background knowledge about the graph convolutions, on which our proposed approach is further built. 

\vspace{0.5em}

\textbf{Notations.} Vectors are denoted by boldface lowercase letters, and matrices are denoted by boldface capital letters. An element of a vector $\mathbf{x}$ is denoted by $x_i$, and an element of a matrix $\mathbf{X}$ is denoted by  $\mathbf{X}_{ij}$. 
For any vector $\mathbf{x} \in \mathbb{R}^n$, $Diag(\mathbf{x}) \in \mathbb{R}^{n \times n}$ is the diagonal matrix whose diagonal elements are $x_i$. $\mathbf{I}_n$ denotes an identity matrix with size $n$. We denote an undirected graph as $G=(V, E, \mathbf{A})$, where $V$ is the set of nodes, $E \subset V \times V $ is the set of edges, and $\mathbf{A} \in \mathbb{R}^{n \times n}$ is the weighted adjacency matrix, where the entry $\mathbf{A}_{ij}$ denotes the pairwise affinity between node $i$ and node $j$ of graph $G$. 

Before formally defining our problem, we first introduce the concept of ``brain network" and ``module" in brain networks. A brain network is a weighted undirected graph $G=(V, E, \mathbf{A})$, where each node $v_i$ in $V$ denotes a specific brain region of interest (ROI) and the edge connecting $v_i$ and $v_j$ represents the connection between region $v_i$ and region $v_j$, whereas the element $\mathbf{A}_{ij}$ in $\mathbf{A}$ denotes the weight of their connection. In functional brain network derived from fMRI, the edges indicate the functional correlations between two brain regions, while in structural brain network derived from DTI, the edges indicate the neural fiber connections between different regions. A module (or community) in brain networks is a subset of nodes that are densely connected to each other while having sparse connections to the nodes in other modules \cite{sporns2013structure}.   
\vspace{0.5em}

\textbf{Problem Definition.} Assume we are given a set of brain network instances $D=\{G_1,G_2,\cdots ,G_N\}$, and each instance ${G}_i$ incorporates a structural brain network ${G}_i^{(s)} = (V^{(s)}, E^{(s)}, \mathbf{A}^{(s)})$, and a functional brain network ${G}_i^{(f)} = (V^{(f)}, E^{(f)}, \mathbf{A}^{(f)})$, where $V^{(s)}$ and $V^{(f)}$ contain the same number of nodes representing the same set of brain regions, $|V^{(s)}| = |V^{(f)}| = n$, $E^{(s)}$ is the set of edges in ${G}_i^{(s)}$ and $E^{(f)}$ is the set of edges in ${G}_i^{(f)}$, $\mathbf{A}^{(s)} \in \mathbb{R}^{n \times n}$ is the adjacency matrix of ${G}_i^{(s)}$, and $\mathbf{A}^{(f)} \in \mathbb{R}^{n \times n}$ is the weighted adjacency matrix of ${G}_i^{(f)}$, where each element represents the functional correlation between two brain regions. We aim to obtain a network embedding $\mathbf{Z} \in \mathbb{R}^{n \times d}$ for each ${G}_i$ by jointly learning from ${G}_i^{(s)}$ and ${G}_i^{(f)}$, where $d$ represents the dimension of each node embedding. The joint embedding should not only capture both the inherent structural information and functional characteristics of the brain network, but should also preserve the underlying community/modular structure of the brain network.

\vspace{0.5em}

In this paper, we will build a novel framework based on graph convolutions for learning the structural and functional joint embedding of brain networks. Applying filters on the spectral domain of structural and functional brain networks will enhance the ability of our model in pruning noisy information and strengthening relevant signals for learning accurate brain network embedding and distinguishing complicated brain disorders.

\section{Framework} \label{sec_framework}
\begin{figure*}[t]
\centering
    \begin{minipage}[l]{\linewidth}
      \centering
      \includegraphics[width=0.70\linewidth]{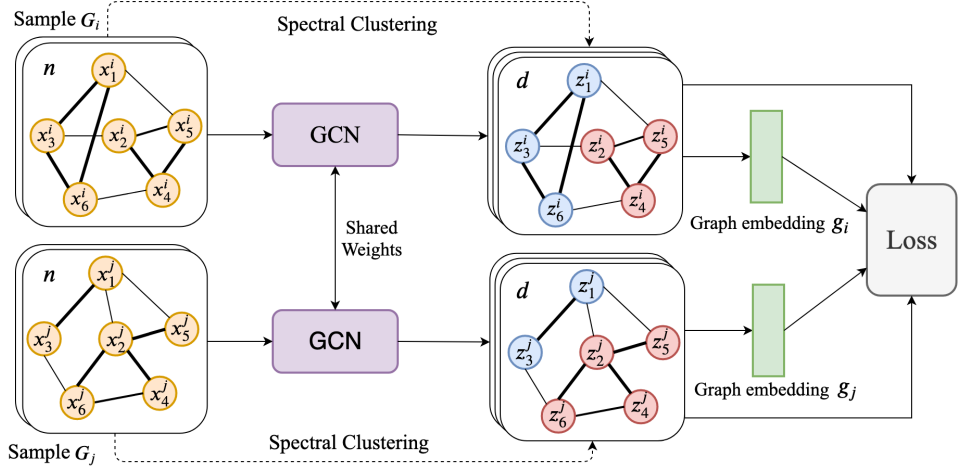}
    \end{minipage}
  \caption{An overview of the proposed SCP-GCN framework. Each of the two input samples is a graph whose nodes represent brain regions and the connections (\ie edges) between these brain regions  are defined by the DTI brain network. An $n$-dimensional feature vector $\mathbf{x}^{i}_w$ is assigned on each node $w$ of sample $G_i$, which is the adjacent vector of the corresponding node in the fMRI functional network. The Siamese GCN takes a pair of samples $<G_i,G_j>$ as input, and learns a $d$-dimensional node embedding $\mathbf{z}^{i}_w$ for each node of $G_i$ and $\mathbf{z}^{j}_w$ for each node of $G_j$, which are then concatenated into a graph embedding $\mathbf{g}_i$ and $\mathbf{g}_j$, respectively. Spectral clustering is employed on the DTI brain network to get the communities of nodes (shown by different node colors in the figure). The graph embeddings, node embeddings, and community information are used to compute the loss function elaborated in section \uppercase\expandafter{\romannumeral3}.}\label{fig:framework}
\end{figure*}

In this section, we introduce the proposed Siamese community-preserving graph convolutional network (SCP-GCN) framework for structural and functional joint embedding of brain networks. We first describe how we use graph convolutions for jointly learning an embedding from structural and functional brain networks. Then we introduce the Siamese GCN model, after which we present how we incorporate the community-preserving property into the framework to enable it to learn an embedding that preserves the community structure in brain networks. An overview of the framework is shown in Fig.~\ref{fig:framework}.

\subsection{Graph Convolutions for Structural and Functional Joint Embedding}\label{sec_gcn}

We propose to use graph convolutions to learn the structural and functional joint embedding of brain networks. In the graph convolutions, the graph structure is defined by the structural connectivity and the node features come from functional connectivity. By considering the structural and functional networks jointly in this way, both the inherent structural information and the functional patterns can be captured during the learning process. 

Given a brain network instance $G$ from $D$, with structural brain network
${G}^{(s)} = (V^{(s)}, E^{(s)}, \mathbf{A}^{(s)})$ and a functional brain network ${G}^{(f)} = (V^{(f)}, E^{(f)}, \mathbf{A}^{(f)})$, we set $\mathbf{A} = \mathbf{A}^{(s)}$ as the adjacency matrix for the graph. Then the normalized graph Laplacian can be defined as $\mathbf{L} = \mathbf{I}_{n} - \mathbf{D}^{-\frac{1}{2}}\mathbf{A}\mathbf{D}^{-\frac{1}{2}}$, where $\mathbf{I}_{n}$ is an identity matrix and $\mathbf{D}\in\mathbb{R}^{n\times n}$ is the diagonal degree matrix of the graph with leading entries $\mathbf{D}_{ii}=\sum_{j}\mathbf{A}_{ij}$. 

Consider a $n$-dimensional signal $\mathbf{x}: V \rightarrow \mathbb{R}^{n}$ defined on graph $G$, which can be regarded as an one-dimensional feature vector, with $x_i \in \mathbb{R}$ assigned to the $i^{th}$ node. According to \cite{shuman2013emerging}, the convolution operation in the Fourier domain can be defined as the multiplication of the signal $\mathbf{x}$ with a filter $g_{\bm{\theta}}$ parameterized by $\bm{\theta} \in \mathbb{R}^{n}$:
\begin{equation}
    g_{\bm{\theta}} * \mathbf{x} = \mathbf{U} g_{\bm{\theta}}(\mathbf{\Lambda}) \mathbf{U}^{T} \mathbf{x}
\label{eq:spectral_filter}
\end{equation}
where $\mathbf{U}=[u_0,\ldots, u_{n-1}]\in\mathbb{R}^{n\times n}$ is the eigenvector matrix of the normalized graph Laplacian $\mathbf{L}$, \ie $\mathbf{L}=\mathbf{U\Lambda U}^{T}$, where $\mathbf{\Lambda} = Diag([\lambda_0,\ldots, \lambda_{n-1}])\in\mathbb{R}^{n\times n}$ is the diagonal matrix of its eigenvalues, and $g_{\bm{\theta}}(\mathbf{\Lambda})=Diag([g_{\bm{\theta}}(\lambda_0),\ldots,g_{\bm{\theta}}(\lambda_{n-1})])$.

To circumvent the expensive computations involved in the multiplication in Equation (\ref{eq:spectral_filter}) and the eigendecomposition of $\mathbf{L}$, we approximate $g_{\bm{\theta}}(\mathbf{\Lambda})$ by a truncated expansion in terms of Chebyshev polynomials $\mathbf{T}_{k}(x)$ up to $K^{th}$ order, as suggested in \cite{hammond2011wavelets}:

\begin{equation}
    g_{\bm{\theta}} * \mathbf{x} \approx \sum_{k=0}^{K}{\theta_k}\mathbf{T}_k(\hat{\mathbf{L}}) \mathbf{x}
\label{eq:K_local}
\end{equation} 
where $\mathbf{\hat{L}} = 2\mathbf{L}/\lambda_{max} - \mathbf{I}_n$, $\lambda_{max}$ denotes the largest eigenvalue of $\mathbf{L}$, and $\bm{\theta} \in \mathbb{R}^{K}$ is now a vector of Chebyshev coefficients. The Chebyshev polynomials are recursively defined as $\mathbf{T}_k(\mathbf{x}) = 2\mathbf{x}\mathbf{T}_{k-1}(\mathbf{x}) - \mathbf{T}_{k-2}(\mathbf{x})$ with $\mathbf{T}_{0}(\mathbf{x}) = 1$ and $\mathbf{T}_{1}(\mathbf{x}) = \mathbf{x}$. As it is a $K^{th}$-order polynomial in the Laplacian, the convolution is now $K$-localized, which means it depends on nodes that are at maximum $K$ steps away from the node \cite{hammond2011wavelets,defferrard2016convolutional}.

According to \cite{kipf2016semi}, we limits the layer-wise convolution operation to $K = 1$ in order to alleviate the potential problem of overfitting on local neighborhood structures for graphs that have wide node degree distributions. By further approximating $\lambda_{max} \approx 2$, Equation (\ref{eq:K_local}) can be simplified to:
\begin{equation}
    g_{\bm{\theta}'} * \mathbf{x} \approx {{\theta}'_0}\mathbf{x} + {{\theta}'_1}(\mathbf{L} - \mathbf{I}_n)\mathbf{x} = {{\theta}'_0}\mathbf{x} - {{\theta}'_1}\mathbf{D}^{-\frac{1}{2}}\mathbf{A}\mathbf{D}^{-\frac{1}{2}}\mathbf{x}
\end{equation}
with two parameters ${{\theta}'_0}$ and ${{\theta}'_1}$. By further constraining the number of parameters with $\theta = {\theta}'_0 = {\theta}'_1$, we have the following equation:
\begin{equation}
    g_{\theta} * \mathbf{x} \approx (\mathbf{I}_n + \mathbf{D}^{-\frac{1}{2}}\mathbf{A}\mathbf{D}^{-\frac{1}{2}})\mathbf{x}
\end{equation}

Then we apply the renormalization trick introduced in \cite{kipf2016semi}: $\mathbf{I}_n + \mathbf{D}^{-\frac{1}{2}}\mathbf{A}\mathbf{D}^{-\frac{1}{2}} \rightarrow \mathbf{\hat{D}}^{-\frac{1}{2}}\mathbf{\hat{A}}\mathbf{\hat{D}}^{-\frac{1}{2}}$, with $\mathbf{\hat{A}} = \mathbf{A} + \mathbf{I}_n$ and $\mathbf{\hat{D}}_{ii}=\sum_{j}\mathbf{\hat{A}}_{ij}$, and we generalize the definition to a signal $\mathbf{X} \in \mathbb{R}^{n\times p}$ with $p$ input channels~\cite{kipf2016semi}. Specifically, in our structural and functional joint embedding scenario, since the functional brain network is derived from the fMRI signals, which capture the brain activity features of each brain region, we propose to use the functional correlation matrix $\mathbf{A}^{(f)}$ as the input signal matrix, \ie $\mathbf{X} = \mathbf{A}^{(f)}$ with the number of input channels $p = n$.

In this paper, we consider a multi-layer graph convolutional network with the convolutions defined above and the layer-wise propagation rule proposed in \cite{kipf2016semi}. Assume the activation of the $l$-th layer is represented as $\mathbf{H}^{(l)}\in\mathbb{R}^{n\times d}$ and the layer-specific trainable weight matrix for the $l$-th layer is denoted by $\mathbf{\Theta}^{(l)}$, according to the propagation rule, we have:
\begin{equation}
    \mathbf{H}^{(l+1)} = \sigma(\mathbf{\tilde{D}}^{-\frac{1}{2}}\mathbf{\tilde{A}}\mathbf{\tilde{D}}^{-\frac{1}{2}}\mathbf{H}^{(l)}\mathbf{\Theta}^{(l)})
\end{equation}
where $\mathbf{\tilde{A}} = \mathbf{A} + \mathbf{I}_{n}$ is the weight matrix of the undirected graph with added self-connections, and $\sigma(\cdot)$ denotes the activation function. $\mathbf{H}^{(0)} = \mathbf{A}^{(f)} =\mathbf{X}$ is the input feature matrix.


In the problem setting, we take the node features in functional brain network as the input features of the graph, and set $\mathbf{H}^{(0)}=\mathbf{A}^{(f)}\in\mathbb{R}^{n\times n}$, which is the weighted adjacent matrix of functional brain network. That is to say, for the $i^{th}$ node $V_i^{(s)} \in V^{(s)}$, we assign $\mathbf{a}_i^{(f)}\in\mathbb{R}^{n}$, the $i^{th}$ row of $\mathbf{A}^{(f)}$, as the feature vector of that node. The output $\mathbf{Z} = \mathbf{H}^{(l)}$ of the last layer will be the final node embedding of the brain network, where the $i^{th}$ row of $\mathbf{Z}$ represent the embedding vector for the $i^{th}$ node. For the further calculation of the distance between graphs for similarity learning in Siamese network, we concatenate all the rows of $\mathbf{Z}$ into a vector $\mathbf{g}$ as the graph embedding for $G$.


\subsection{Siamese Graph Convolutional Network} \label{sec_sgcn}

Siamese network is first proposed in \cite{bromley1994signature} to solve signature verification as an image matching problem. The Siamese network takes a pair of inputs, and output the similarity between the inputs. \cite{koch2015siamese} introduces this architecture into one-shot learning problem setting in which correct predictions must be given only based on a single training sample of each new class, demonstrating the superior learning ability enabled by Siamese network, even with a small sample size. 

In this paper, we use the Siamese architecture and exploit the pair-wise similarity learning of brain networks to guide the learning process, and also to help address the data scarcity problem caused by the limited sample size.

Contrastive loss function \cite{koch2015siamese} is used to train Siamese network: 
\begin{equation}
    L_S = \frac{y}{2}\|\mathbf{g}_i - \mathbf{g}_j\|_2^2 + (1 - y)\frac{1}{2}\{max(0, m- \| \mathbf{g}_i - \mathbf{g}_j\|_2)\}^2
\label{eq:siames}    
\end{equation}
where $\mathbf{g}_i$ and $\mathbf{g}_j$ are the graph embeddings of instance $i$ and $j$ computed from the GCN, $m$ is a margin value which is greater than $0$. $y=1$ if two input sample are from the same class and $y=0$ if they are from the different classes. The loss function minimizes the Euclidean distance between two input vectors when two samples are from the same class, and maximizes it when they belong to different classes. 



\subsection{Siamese Community-preserving GCN Framework}
In order to preserve the community structure in brain networks, we propose to incorporate the community-preserving property into the Siamese GCN model. We integrate the community-preserving property and the pair-wise similarity learning strategy into a unified framework and call it Siamese Community-Preserving GCN (SCP-GCN). 

The goal of community preserving is that if two nodes are from the same community in the original graph, the Euclidean distance between the learned node embeddings should be small, and if the two nodes are from different communities in the original graph they should have a large distance in their node embedding space. As shown in Fig.~\ref{fig:framework}, we employ spectral clustering \cite{von2007tutorial} to detect the communities from the original structural network. Spectral clustering is an approach for identifying communities of nodes in a graph based on the eigenvalues (spectrum) of Laplacian matrix built from the graph, and it has been shown to be an effective way to obtain the community/modular structure in brain networks \cite{ma2016multi,cribben2017estimating}. Therefore, we employ the spectral clustering algorithm \cite{shi2000normalized} on ${G}^{(s)}$ to capture the community structure of the structural brain network in the original space, and we aim to preserve this community structure in the learning process of Siamese GCN. 

Each community $c$ detected in the network ${G}^{(s)}$ is represented as a set $S_c$, which contains the indexes of nodes belonging to community $c$. We compute a community center embedding $\hat{\mathbf{z}}_c = \frac{1}{|S_c|}\sum_{i \in S_c}\mathbf{z}_i$ for each community $c$, where $\mathbf{z}_i$ is the embedding of the $i^{th}$ node, i.e., the $i^{th}$ row in the graph embedding $\mathbf{Z}$.

The community-preserving objective consists of two components: 1) minimizing the intra-community loss, i.e. the distance between community center $\hat{\mathbf{z}}_c$ and node embeddings belonging to community, and 2) maximizing the inter-community loss, i.e., the distance between the centers of different communities. By combining these two parts, we have the following function as community-preserving loss:


\begin{equation} 
\label{eq:community_loss}
    L_{CP} = \alpha (\sum_{c}\frac{1}{|S_{c}|}\sum_{i \in S_{c}} \| \mathbf{z}_i - \hat{\mathbf{z}}_c\|_2^2) - \beta \sum_{c, c'} \| \hat{\mathbf{z}}_c - \hat{\mathbf{z}}_{c'} \|_2^2
\end{equation} 
where the first part computes the Euclidean distance between node embedding $\mathbf{z}_i$ and its community center $\hat{\mathbf{z}}_c$. The second part computes the Euclidean distance between community center $\hat{\mathbf{z}}_c$ and $\hat{\mathbf{z}}_{c'}$. $\alpha$ and $\beta$ are weights of intra/inter-community loss.

Now we have the overall loss function for our SCP-GCN framework, which can be written as 
\begin{equation}
    L = \sum_{i,j} L_S + \sum_{i}^{N} L_{CP}
\label{eq:total_loss}
\end{equation}
By combining the contrastive loss with community-preserving loss in this way, we can leverage the community structure in the learning process of GCN in Siamese network. The community structure in the original brain network could be preserved in the embedding space when we minimize the loss in Equation (\ref{eq:total_loss}). After the training process, we can use either branch of the twin GCN networks in SCP-GCN for computing a structural and functional joint graph embedding for a given brain network, and the output graph embedding will not only contain group-contrasting features, but also preserve the community structure of the structural brain network, both of which are important for further neurological disorder analysis.

\section{Experiments} \label{sec_experiment}
In order to evaluate the proposed framework for structural and functional joint embedding of brain networks, we compare our approach with the state-of-the-art methods in this field on two real-world brain datasets for neurological analysis. In this section, we first introduce the datasets used in the experiments, the compared baseline methods and experiment settings. Next, we show the performance of all the methods on the two brain network datasets. Then we further present some case studies about the structural and functional brain network analysis, as well as the community structure of brain networks.    
\subsection{Datasets and Preprocessing}
In this work, we use two real datasets as follows:
\begin{itemize}
\item \textit{Human Immunodeficiency Virus Infection (HIV)}: This dataset is collected from the Chicago Early HIV Infection Study at Northwestern University\cite{ragin2012structural}. This clinical study involves 77 subjects, 56 of which are early HIV patients (positive)  and the other 21 subjects are seronegative controls (negative). These two groups of subjects do not differ in demographic characteristics such as age, gender, racial composition and education level. This dataset contains both the functional magnetic resonance imaging (fMRI) and diffusion tensor imaging (DTI) for each subject, from which we can construct the fMRI and DTI brain networks. Then we can treat them as graphs with two views.
\item \textit{Bipolar}: This dataset consists of the fMRI and DTI image data of 52 bipolar I subjects who are in euthymia and 45 healthy controls with matched age and gender. The resting-state fMRI scan was acquired on a 3T Siemens Trio scanner using a T2*-weighted echo planar imaging (EPI) gradient-echo pulse sequence with integrated parallel acquisition technique (IPAT) and diffusion weighted MRI data were acquired on a Siemens 3T Trio scanner. The detailed description of the dataset can be found in \cite{cao2015identification}.
\end{itemize}

We perform preprocessing on the HIV dataset using the standard process as illustrated in \cite{cao2015identifying}. First, we use the DPARSF toolbox\footnote{http://rfmri.org/DPARSF.} to process the fMRI data. We realign the images to the first volume, do the slice timing correction and normalization, and then use an 8-mm Gaussian kernel to smooth the image spatially. The band-pass filtering ($0.01$-$0.08$ Hz) and linear trend removing of the time series are also performed. We focus on the 116 anatomical volumes of interest (AVOI), each of which represents a specific brain region, and extract a sequence of responds from them. Finally, we construct a brain network with the 90 cerebral regions. Each node in the graph represents a brain region and links are created based on the correlations between different brain regions. For DTI data, we use FSL toolbox\footnote{http://fsl.fmrib.ox.ac.uk/fsl/fslwiki.} for preprocessing and then construct the brain networks. The preprocessing includes distortion correction, noise filtering, repetitive sampling from the distributions of principal diffusion directions for each voxel. We parcellate the DTI images into the 90 regions same with fMRI via the propagation of the Automated Anatomical Labeling (AAL) on each DTI image \cite{tzourio2002automated}.

For the Bipolar dataset, the brain networks were constructed using the CONN\footnote{http://www.nitrc.org/projects/conn} toolbox \cite{whitfield2012conn}. The raw EPI images were first realigned and co-registered, after which we perform the normalization and smoothing. Then the confound effects from motion artifact, white matter, and CSF were regressed out of the signal. Finally, the brain networks were derived using the pairwise signal correlations based on the 82 labeled Freesurfer-generated cortical/subcortical gray matter regions.

\begin{table}[t]
\small
\caption{Classification Accuracy (mean $\pm$ std).}
\label{tab:results_acc}
\centering
\resizebox{0.9\columnwidth}{!}{
\begin{tabular}{lcc}
\toprule
Methods        &\emph{Bipolar}   &\emph{HIV}\\
\midrule
DeepWalk
&$0.520 \pm 0.034$      &$0.575 \pm 0.041$\\
node2vec
&$0.555 \pm 0.031$      &$0.625 \pm 0.029$\\
SDBN
&$0.648 \pm 0.010$      &$0.665 \pm 0.010$\\
MVGE-HD
&$0.656 \pm 0.012$      &$0.681 \pm 0.015$\\
GCN
&$0.547 \pm 0.038$      &$0.618 \pm 0.049$\\
CP-GCN
&$0.562 \pm 0.039$      &$0.648 \pm 0.061$\\
S-GCN
&$0.649 \pm 0.033$      &$0.701 \pm 0.090$\\
SCP-GCN
&\textbf{0.677} $\pm$ \textbf{0.033}  &\textbf{0.768} $\pm$ \textbf{0.110}\\
\bottomrule
\end{tabular}}
\end{table}

\begin{table}[t]
\small
\caption{Classification F1 score (mean $\pm$ std).}
\label{tab:results_F1}
\centering
\resizebox{0.9\columnwidth}{!}{
\begin{tabular}{lcc}
\toprule
Methods        &\emph{Bipolar}   &\emph{HIV}\\
\midrule
DeepWalk
&$0.589 \pm 0.024$      &$0.634 \pm 0.021$\\
node2vec
&$0.614 \pm 0.029$      &$0.640 \pm 0.015$\\
SDBN
&$0.637 \pm 0.010$      &$0.667 \pm 0.010$\\
MVGE-HD
&$0.661 \pm 0.010$      &$0.705 \pm 0.011$\\
GCN
&$0.612 \pm 0.029$      &$0.713 \pm 0.039$\\
CP-GCN
&$0.617 \pm 0.036$      &$0.766 \pm 0.055$\\
S-GCN
&$0.744 \pm 0.029$      &$0.787 \pm 0.010$\\
SCP-GCN
&\textbf{0.750} $\pm$ \textbf{0.033}  &\textbf{0.840} $\pm$ \textbf{0.010}\\
\bottomrule
\end{tabular}}
\end{table}

\subsection{Baselines and Metrics}\label{sec_baselines}
We compare the proposed SCP-GCN framework with other state-of-the-art methods in structural and functional embedding of brain network for neurological disorder analysis. We also compare with several variations of the proposed approach to evaluate the importance of the different key components in SCP-GCN. All the compared methods are summarized as below.
\begin{itemize}
    \item \textbf{DeepWalk}: DeepWalk \cite{deepwalk} is a method for learning node embedding in graphs. It uses local information obtained from random walks on graphs to learn latent node representations. In our experiments, we run DeepWalk on DTI and fMRI networks separately to get the node embedding of each network. Graph embedding is obtained by concatenating the node embeddings of the two networks.
    \item \textbf{node2vec}: node2vec \cite{node2vec} learns node embedding by extending DeepWalk with more complicated random walk or search method. In our experiment, we follow the same experiment setup as DeepWalk.
    \item \textbf{SDBN} \cite{wang2017structural}: It is a CNN based deep learning method, which uses graph reordering and structural augmentation to capture the structural information in brain networks and learns a feature representation for brain disorder classification. Since this method can only deal with single-view brain network, we apply it on fMRI brain network and DTI brain network respectively and report the best performance from the two cases.
    \item \textbf{MVGE-HD}: It is a multi-view  graph  embedding method proposed in \cite{ma2017multiview} for jointly learning multi-view embedding and hubs from brain networks. In the evaluation, we treat fMRI brain network and DTI brain networks as two views and apply the MVGE-HD to get the embedding of all the instances. 
    \item \textbf{GCN}: It is the graph convolutional network approach presented in \cite{kipf2016semi}. We apply it on the fMRI and DTI brain networks by using DTI structural connectivity as graph structure and using fMRI functional connectivity as node features during the graph convolutions to learn the structural and functional joint embedding.
    \item \textbf{CP-GCN}: It is the graph convolutional network approach with the community-preserving property, i.e., the proposed SCP-GCN without Siamese architecture.
    \item \textbf{S-GCN}: It is the Siamese graph convolutional network introduced in \ref{sec_sgcn}, i.e., the proposed SCP-GCN without community-preserving property. 
    \item \textbf{SCP-GCN}: It is the full framework proposed in this paper, i.e., the Siamese community-preserving graph convolutional network. 
\end{itemize}

To evaluate the quality of the learned brain network embedding for neurological disorder analysis, we feed the learned brain network representation to a sigmoid classifier for neurological disorder detection. We use accuracy and F1 score as the evaluation metrics. We run each experiment for 100 times and report the average performance.

\subsection{Experimental Setup}
In the experiments, we randomly split the subjects into training set and testing set, with $60\%$ for training and the rest $40\%$ for testing. For all the GCNs in the compared methods, we use $2$ convolutional layers followed with $1$ fully connected layer, with $256$ features for the first convolutional layer and $128$ features for the second convolutional layer. We use binary cross entropy loss \cite{robert2014machine} for the baseline GCN method. To train SCP-GCN, we first prepare pairs with the subjects in the training set, and set label $1$ for each pair of the same class and label $0$ for each pair of different classes. For example, given a pair of subjects in Bipolar dataset, if the two subjects are both Bipolar patients or both normal controls, the label for this pair will be $1$, otherwise the label will be $0$. We use $m = 0.5$ for the margin value in Equation (\ref{eq:siames}). We use the stochastic gradient descent with Adaptive Moment Estimation (ADAM) optimizer \cite{kingma2014adam} as the optimization algorithm with learning rate $0.01$. The parameters in the neural networks are tuned with 3-fold cross-validation. For the parameter $\alpha$ and $\beta$ in the proposed SCP-GCN model, we do grid search in $\{ {{10^{ - 3}}, \cdots {10^3}}\}$ to find the initial optimal values from this range and then do a further dense  search with smaller step size to find the optimal values. The details will be desribed in~\ref{sec_para}. The optimal value for community number $c$ is selected by the grid search from $\{ 2,3, \cdots ,10\}$. After the training stage of the Siamese network, we use the well-trained GCN in either of the twin networks to obtain the embedding for all the subjects. For the baseline methods proposed in other works, we follow the guidelines they provide and use the optimal parameter configuration for them. 

\subsection{Evaluation Results}
In this section, we evaluate the effectiveness of the learned structural and functional brain network embedding for neurological disorder detection. The average classification accuracy and F1 score are shown in Table~\ref{tab:results_acc} and Table~\ref{tab:results_F1} respectively.

As we can see from Table ~\ref{tab:results_acc} and Table ~\ref{tab:results_F1}, the embedding obtained by the proposed SCP-GCN results in the best performance on both datasets in terms of classification accuracy and F1 score. Among the eight methods, we observe that DeepWalk and node2vec achieve lower accuracy and F1 score compared to the other six methods which are all deep learning models. This indicates that the deep neural networks could better capture the complicated graph features from brain networks for the classification task compared to these two traditional network embedding methods. In addition, SDBN, MVGE-HD, CP-GCN and SCP-GCN are the ones that consider community structure of brain networks during the representation learning. By comparing CP-GCN with GCN and SCP-GCN with S-GCN, we find that adding the community-preserving property helps improve the learning performance of GCN and S-GCN, indicating the importance of community structure in brain network analysis. By comparing S-GCN with GCN and SCP-GCN with CP-GCN, we can see that the pair-wise similarity learning enabled by Siamese network leads to a better learning performance, which shows the pair-wise similarity learning component can guide the representation learning towards a better network embedding for group-contrasting analysis. It can also help reduce the possible over-fitting problem due to the small sample size of brain network data. 

It is worth to mention that, although SDBN is applied on single views, it achieves better results than GCN and CP-GCN, which utilize both structural view and functional view. This is probably because it considers the community structure information and the small-scale setting by augmenting CNN with decoding pathways for reconstruction. In the evaluation, the best performance of SDBN is achieved on fMRI for both Bipolar and HIV datasets, which means the functional brain network provides more discriminative information for the feature learning in SDBN. 

Meanwhile, MVGE-HD, as a multi-view graph embedding method, combines the fMRI view and the DTI view by a weighted sum in their objective function. Although it does not differentiate the fMRI and DTI based on their physical meanings, it achieves a relatively good results, especially when compared to GCN. This is mainly due to the fact that the MVGE-HD approach considers the underlying community structure and hubs in brain networks, and the hub detection component further improves the community-preserving property in the multi-view embedding. Based on these observations, we find that the community structure preserving, structural and functional information integration as well as the Siamese similarity learning are three key factors that facilitate the learning ability of the proposed SCP-GCN approach, resulting in a superior performance of SCP-GCN in neurological disorder detection.  
\begin{figure}[t]
\centering
    \begin{subfigure}[\emph{Accuracy}]{		\includegraphics[width=.46\linewidth] {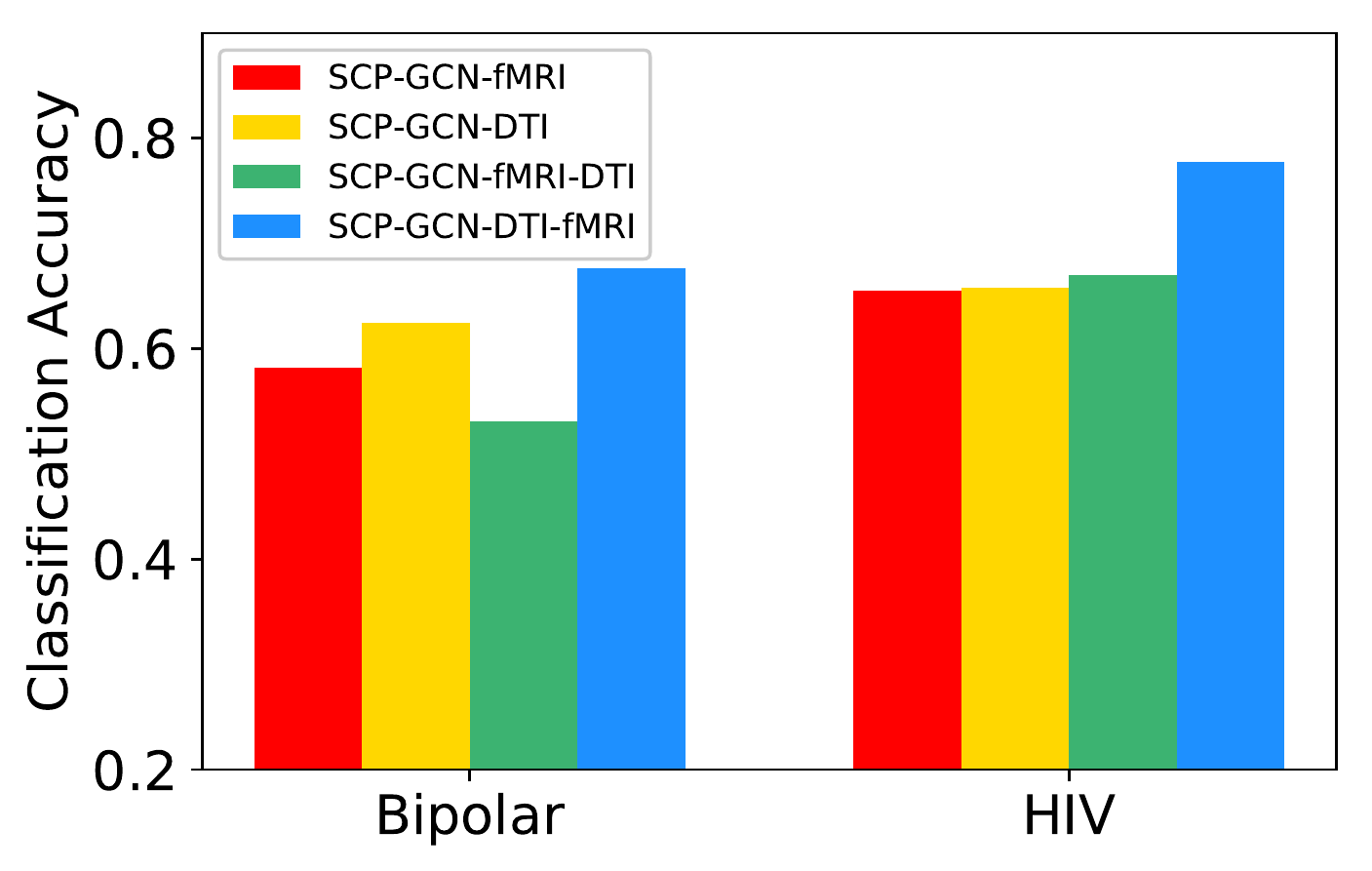}
		    \label{fig:acc_sf}
		}%
		\end{subfigure}
		\begin{subfigure}[\emph{F1 Score}]{
		\includegraphics[width=.46\linewidth]{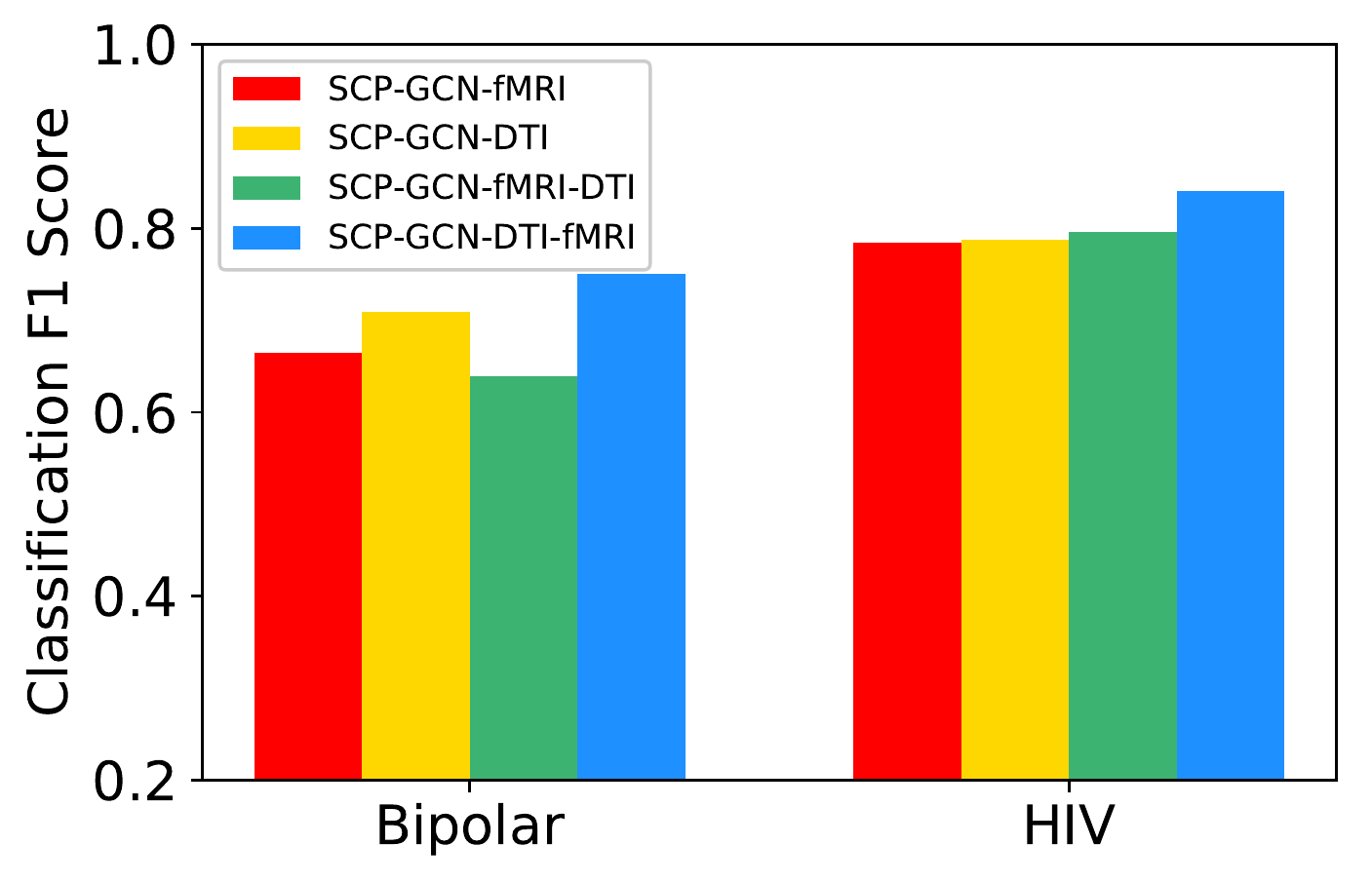}
		\label{fig:f1_sf}
		}%
		\end{subfigure}  
  \caption{Classification Performance in Case Study of Structural and Functional Joint Learning}\label{fig:case1}
\end{figure}

\subsection{Case Studies and Discussions}
In this section, we will provide extensive qualitative analysis and discussion on the structural and functional joint learning and the community-structure preserving property for brain disorder analysis. 
\subsubsection{Case Study of Structural and Functional Joint Learning}\label{sec_casestudy1}
In order to investigate the importance of structural connectivity and functional connectivity in learning brain network embedding, we evaluate and compare the performance of the following variations of the proposed SCP-GCN approach for brain disorder detection on Bipolar and HIV.
\begin{itemize}
    \item \textbf{SCP-GCN-fMRI}: It uses the fMRI functional connectivity for both the graph structure and the node features to be used in the graph convolutions. 
    \item \textbf{SCP-GCN-DTI}: It uses the DTI structural connectivity for both the graph structure and the node features used in the graph convolutions.
    \item \textbf{SCP-GCN-fMRI-DTI}: It uses fMRI functional connectivity as the graph structure while using DTI connectivity for node features.
    \item \textbf{SCP-GCN-DTI-fMRI}: It is the proposed SCP-GCN, where DTI is used to define graph structure and fMRI network is used for node features. 
\end{itemize}


\begin{figure}[t]
\centering
    \begin{subfigure}[\emph{Accuracy}]{		\includegraphics[width=.46\linewidth] {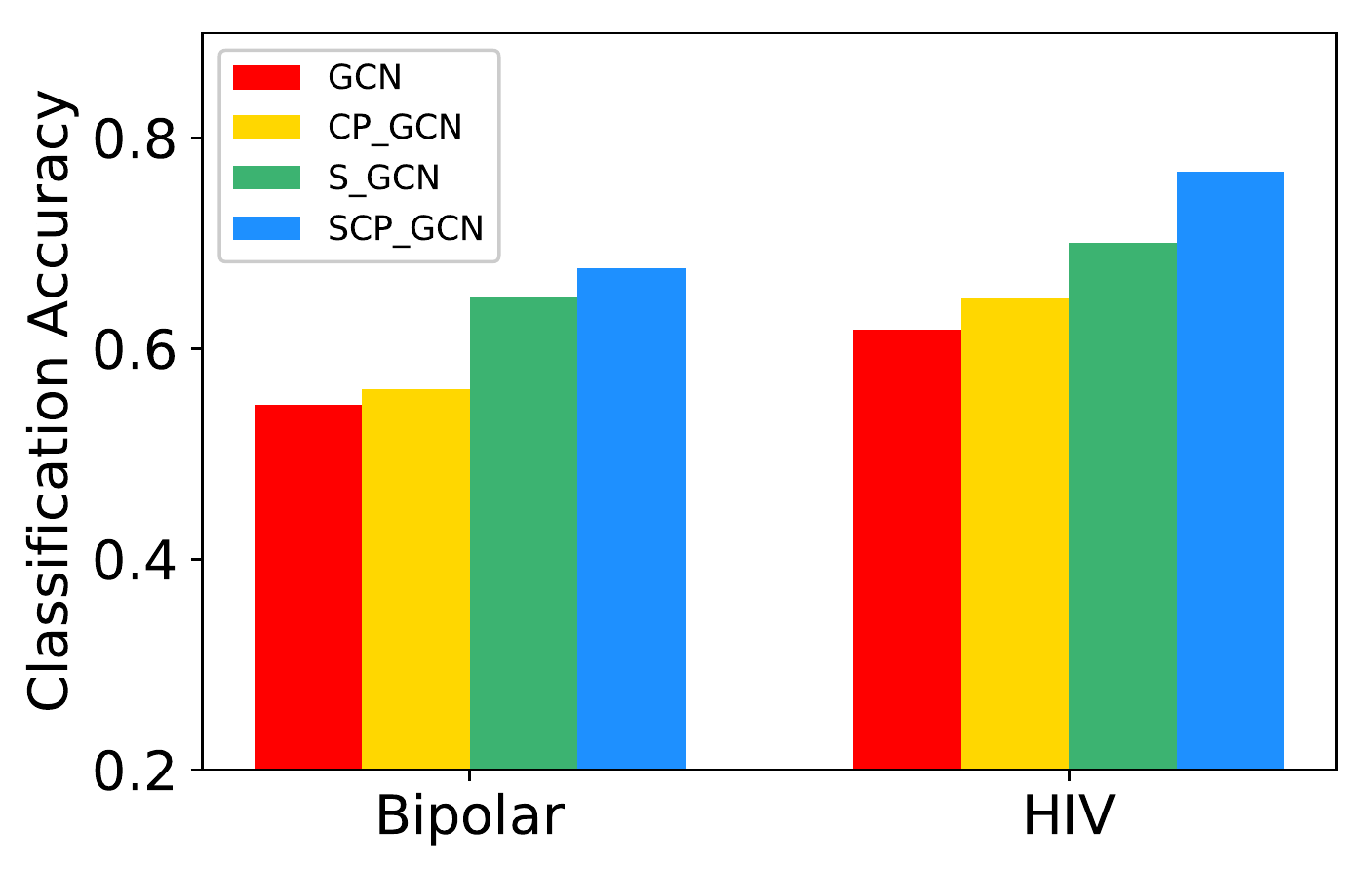}
		    \label{fig:acc_CP}
		}%
		\end{subfigure}
		\begin{subfigure}[\emph{F1 Score}]{
		\includegraphics[width=.46\linewidth]{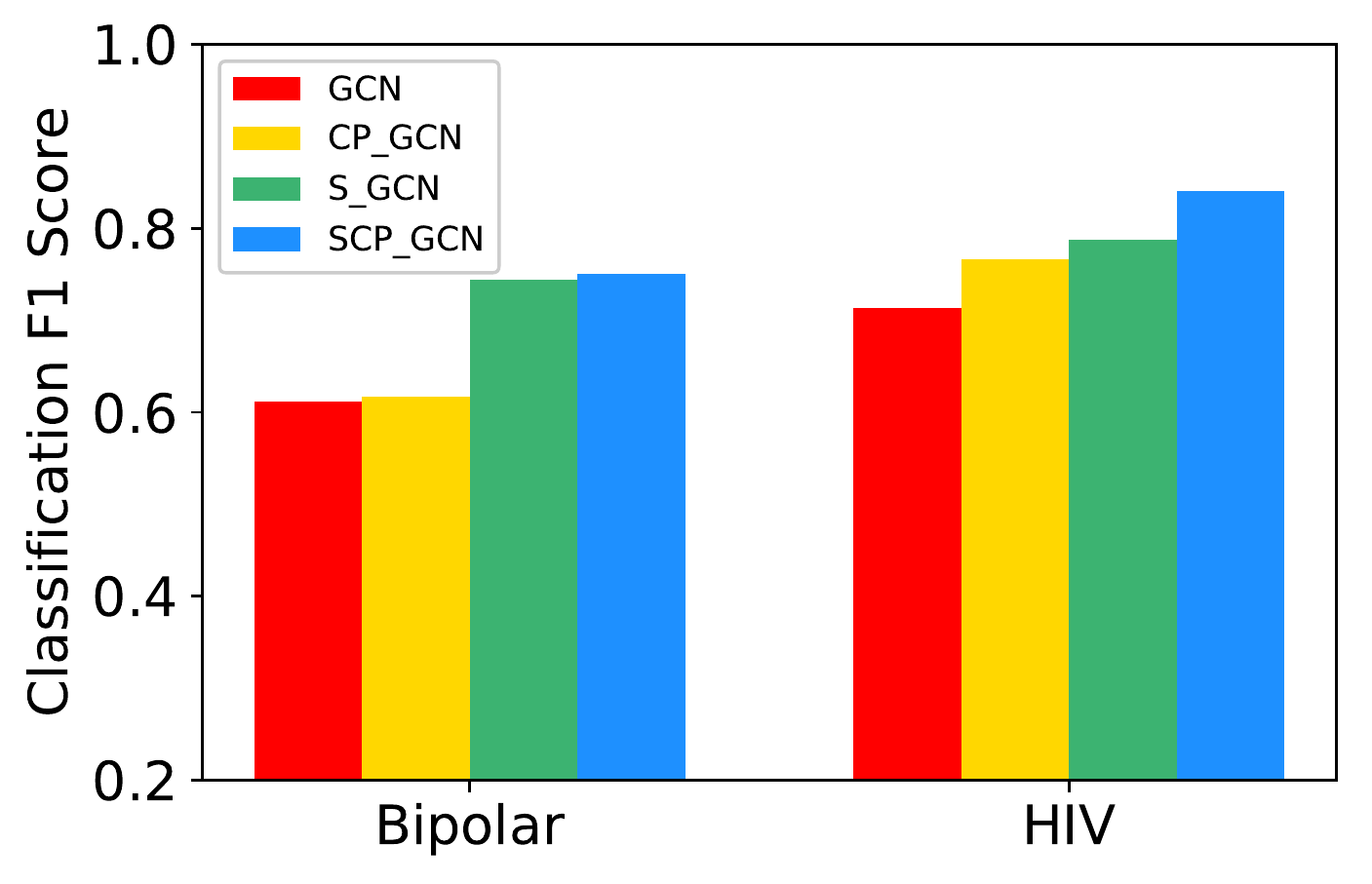}
		\label{fig:f1_CP}
		}%
		\end{subfigure}  
  \caption{Classification Performance in Case Study of Community-Structure Preserving}\label{fig:case2}
\end{figure}
The classification performance of the above four variations in neurological disorder detection on Bipolar and HIV datasets are shown in Fig.~\ref{fig:case1}. From this figure, we can observe that the proposed SCP-GCN-DTI-fMRI version works the best and it greatly outforms the other variations in terms of both accuracy and F1 score. This indicates that using DTI (i.e, structural networks) to define graph structure and using the fMRI (i.e., functional networks) to derive node features works the best in the graph convolutions for group-contrasting brain network embedding analysis. This is reasonable, because DTI reflects the structural connections between brain regions while fMRI reflects the correlations of the brain functional activity patterns of different regions. By using fMRI and DTI jointly in this way, the graph convolutions are performed based on the neighborhood structure reflected by DTI while the node features originating from fMRI are updated based on the neighborhood information during the convolutions. 

If the fMRI and DTI are used vice visa (as in SCP-GCN-fMRI-DTI), the graph structure is determined by the functional connectivity while the node features come from structural connectivity will be updated during convolutions. In this case, the improperly defined graph structure will probably mislead the graph convolutions to use undesirable neighborhood structure for updating node features and the features of functional activity in fMRI will be ignored. The green bars in Fig.~\ref{fig:case1} shows that SCP-GCN-fMRI-DTI method achieves a much lower accuracy and F1 score compared to SCP-GCN-DTI-fMRI, and the accuracy on Bipolar is even lower than the cases using DTI or fMRI only. As shown with the red bars and yellow bars in Fig.~\ref{fig:acc_sf}, the SCP-GCN-fMRI and SCP-GCN-DTI both achieve an accuracy around $60\%$, which is much lower than the accuracy of the proposed SCP-GCN-DTI-fMRI. 

\subsubsection{Case Study of Community-Structure Preserving} \label{sec_casestudy2}
In order to study the importance of the community-preserving property, we compare the performance of the graph convolutions with and without community-preserving property by comparing the following four variations: GCN, CP-GCN, S-GCN, SCP-GCN, which are all described above in Section~\ref{sec_baselines}. As shown in Fig.~\ref{fig:acc_CP} and Fig.~\ref{fig:f1_CP}, compared to GCN, CP-GCN which considers community structure achieves higher accuracy and F1 score on both datasets, and compared to S-GCN, the proposed SCP-GCN performs better, especially in F1 score, which outperforms S-GCN by over $6\%$ on both Bipolar and HIV. This indicates that the community-preserving graph convolutions can learn a more discriminating network representation from structural and functional brain networks for neurological disorder analysis.  

\begin{figure}[t]
\centering
    \begin{subfigure}[\emph{Bipolar}]{		\includegraphics[width=.45\linewidth] {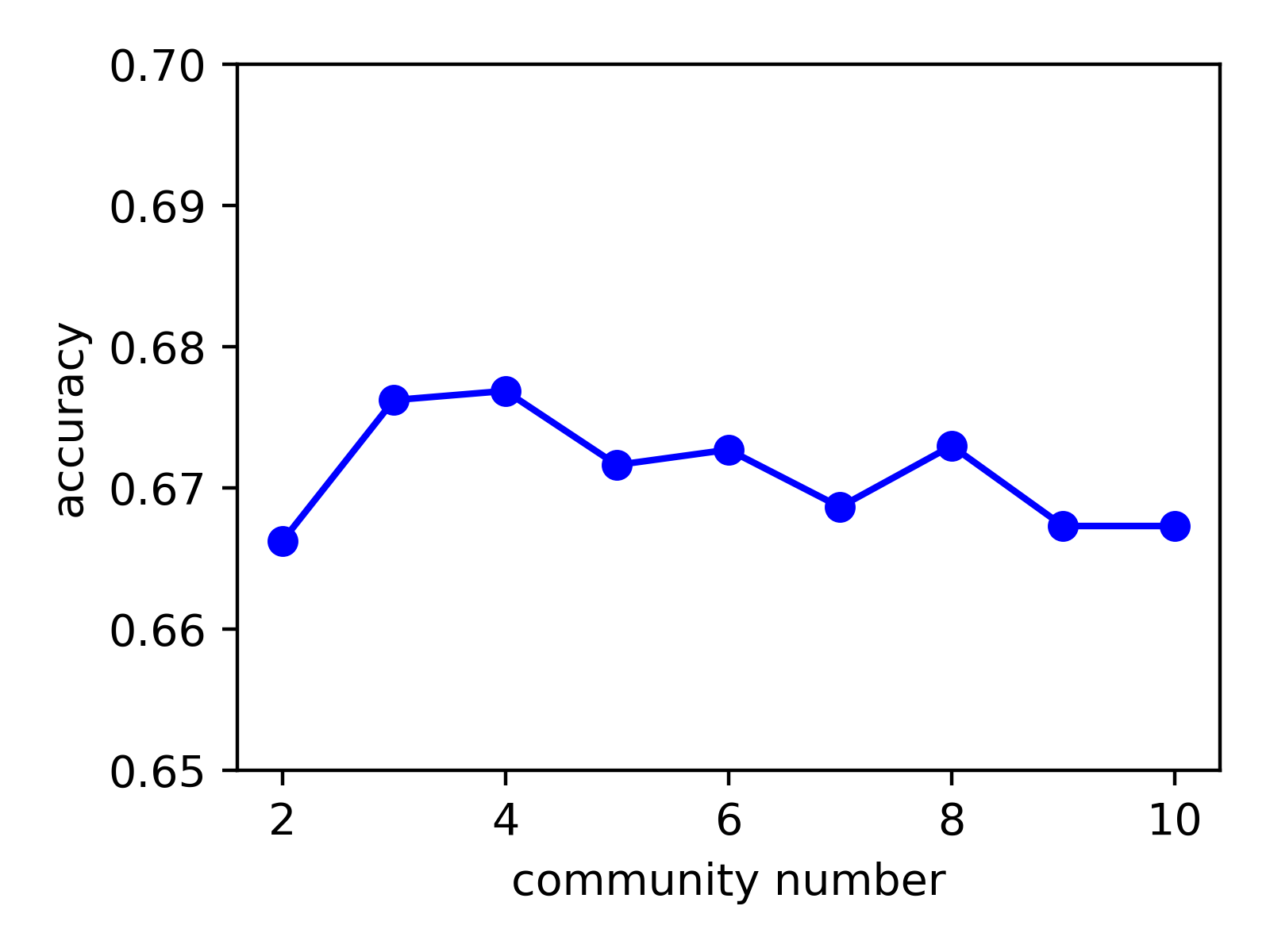}
		    \label{fig:para_c_bipolar}
		}%
		\end{subfigure}
		\begin{subfigure}[\emph{HIV}]{
		\includegraphics[width=.45\linewidth]{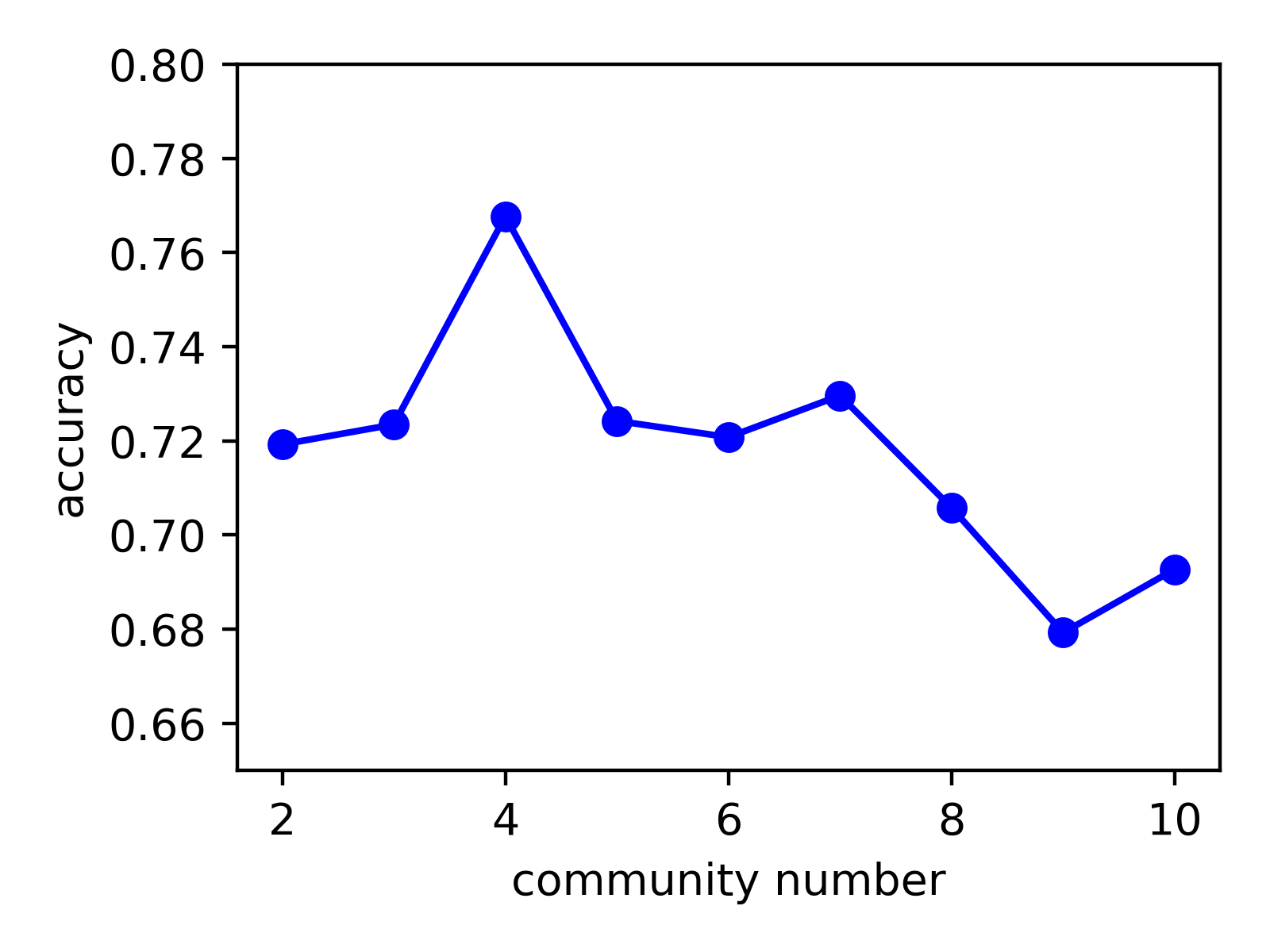}
		\label{fig:para_c_hiv}
		}%
		\end{subfigure}  
  \caption{\emph{Accuracy} with different community numbers}\label{fig:para_c}
\end{figure}

\subsection{Parameter Sensitivity Analysis} \label{sec_para}
In this section, we explore the impact and sensitivity of the three main parameters in the proposed framework, including $\alpha$, $\beta$, which are the weight parameters balancing the trade-off between intra-community loss and inter-community loss in Equation (\ref{eq:community_loss}), as well as the community number $C$. 

Fig.~\ref{fig:para_c} shows the accuracy with different community numbers, where we vary the value of $C$ from 2 to 10. From the figure, we find that the value of $C$ affects the performance of SCP-GCN in classification accuracy. The highest accuracy is achieved when $C = 4$ for both Bipolar and HIV. As we can see from Fig.~\ref{fig:para_c_bipolar} and Fig.~\ref{fig:para_c_hiv}, With the increase of the $C$ value, the performance of SCP-GCN first keeps rising up until it reaches the peak, and then it starts to decline. This changing trend is reasonable as when the community number is very small, there is not enough communities to encode the inter-community relationship among nodes in brain networks, whereas when the community number is too large, it may lose some important intra-community information between nodes. It is worth to mention that, some previous studies in complex brain network analysis identifies four significant modules in human brain \cite{sporns2013structure}, which is consistent with our optimal case, indicating the potential value of our work in neuroscience. 

\begin{figure}[t]
\centering
    \begin{subfigure}[\emph{Bipolar}]{		\includegraphics[width=.46\linewidth] {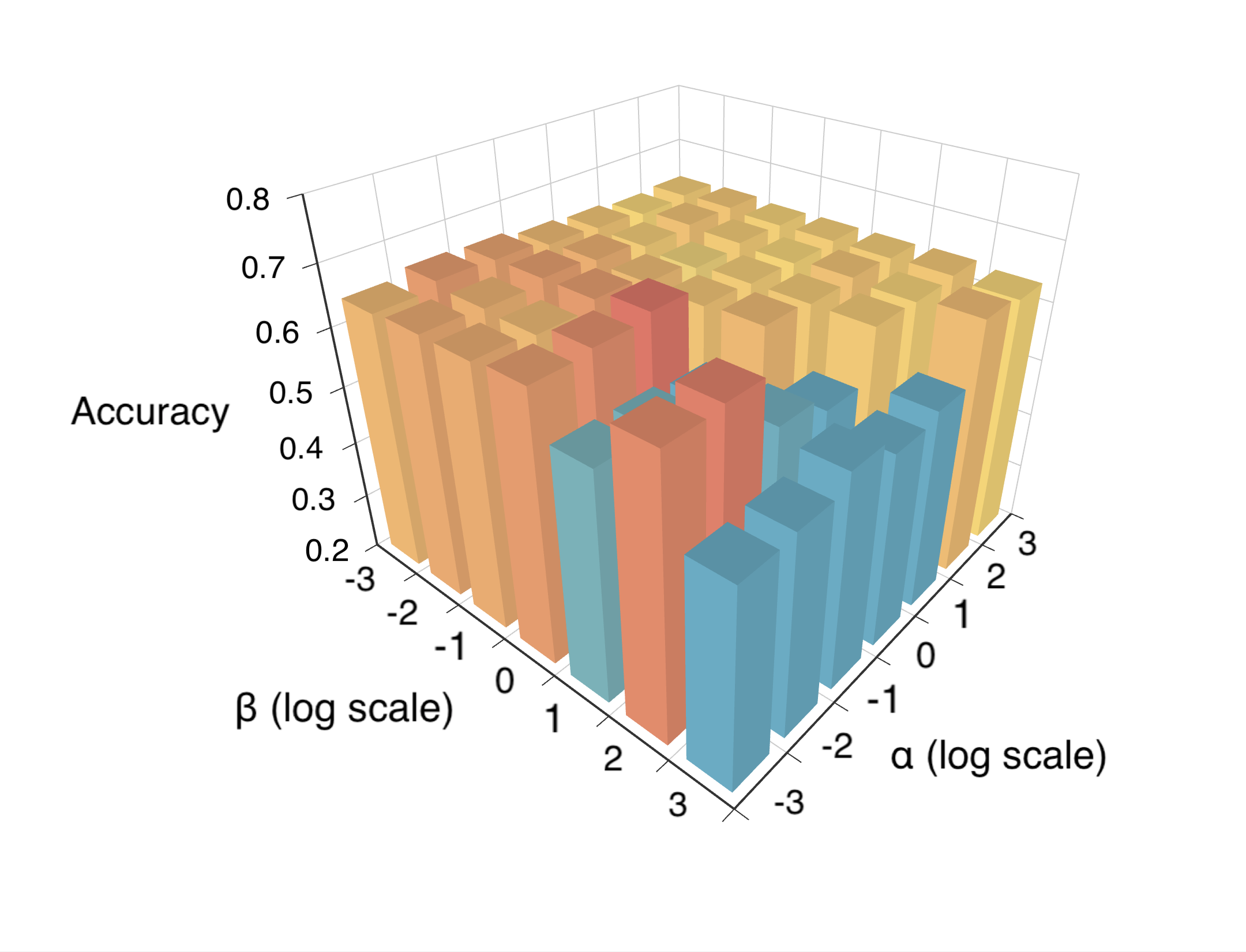}
		    \label{fig:para_bipolar}
		}%
		\end{subfigure}
		\begin{subfigure}[\emph{HIV}]{
		\includegraphics[width=.46\linewidth]{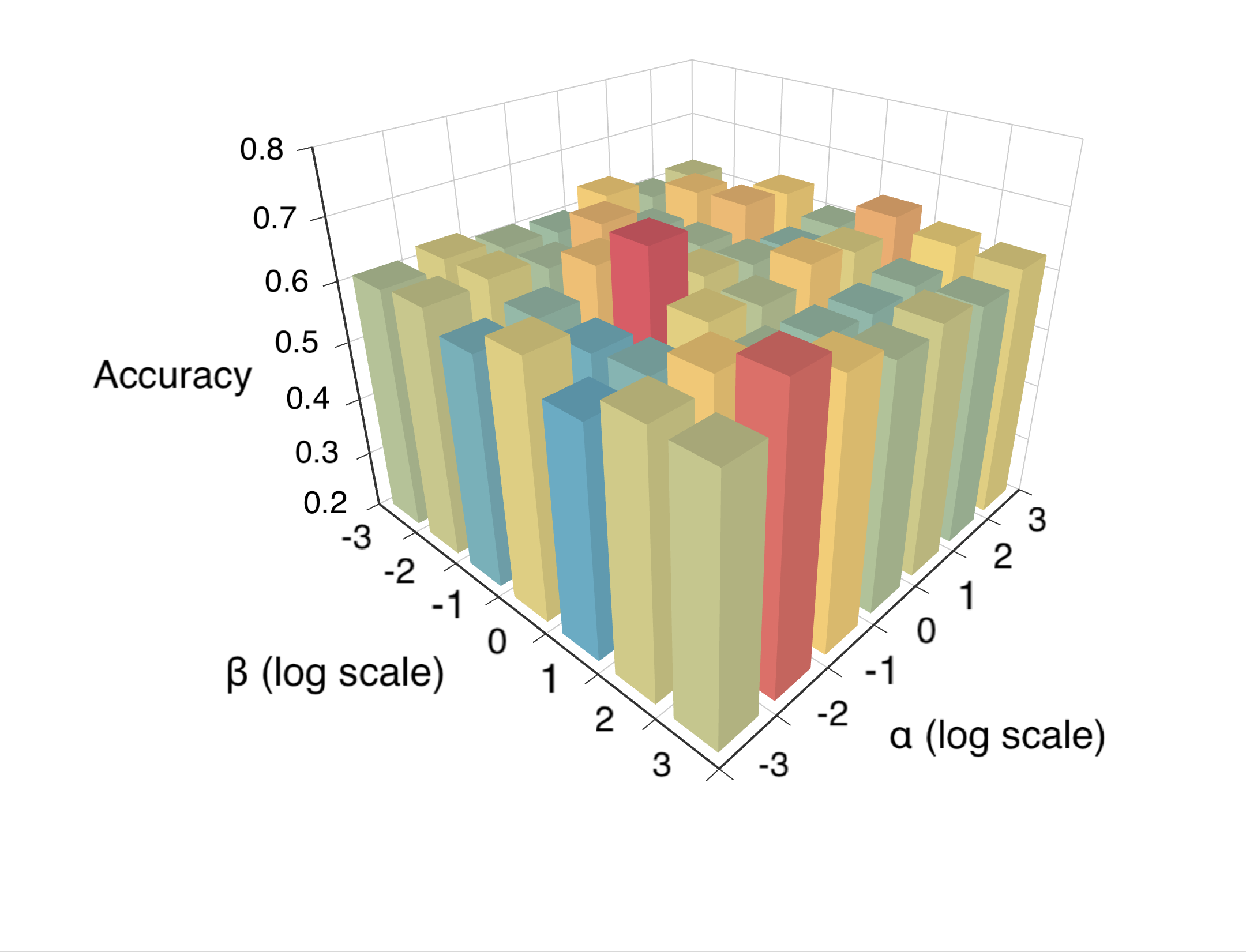}
		\label{fig:para_hiv}
		}%
		\end{subfigure}  
  \caption{\emph{Accuracy} with different values for $\alpha$ and $\beta$}\label{fig:para}
\end{figure}

\begin{figure}
\centering
    \begin{subfigure}[\emph{Bipolar}]{		\includegraphics[width=.75\linewidth] {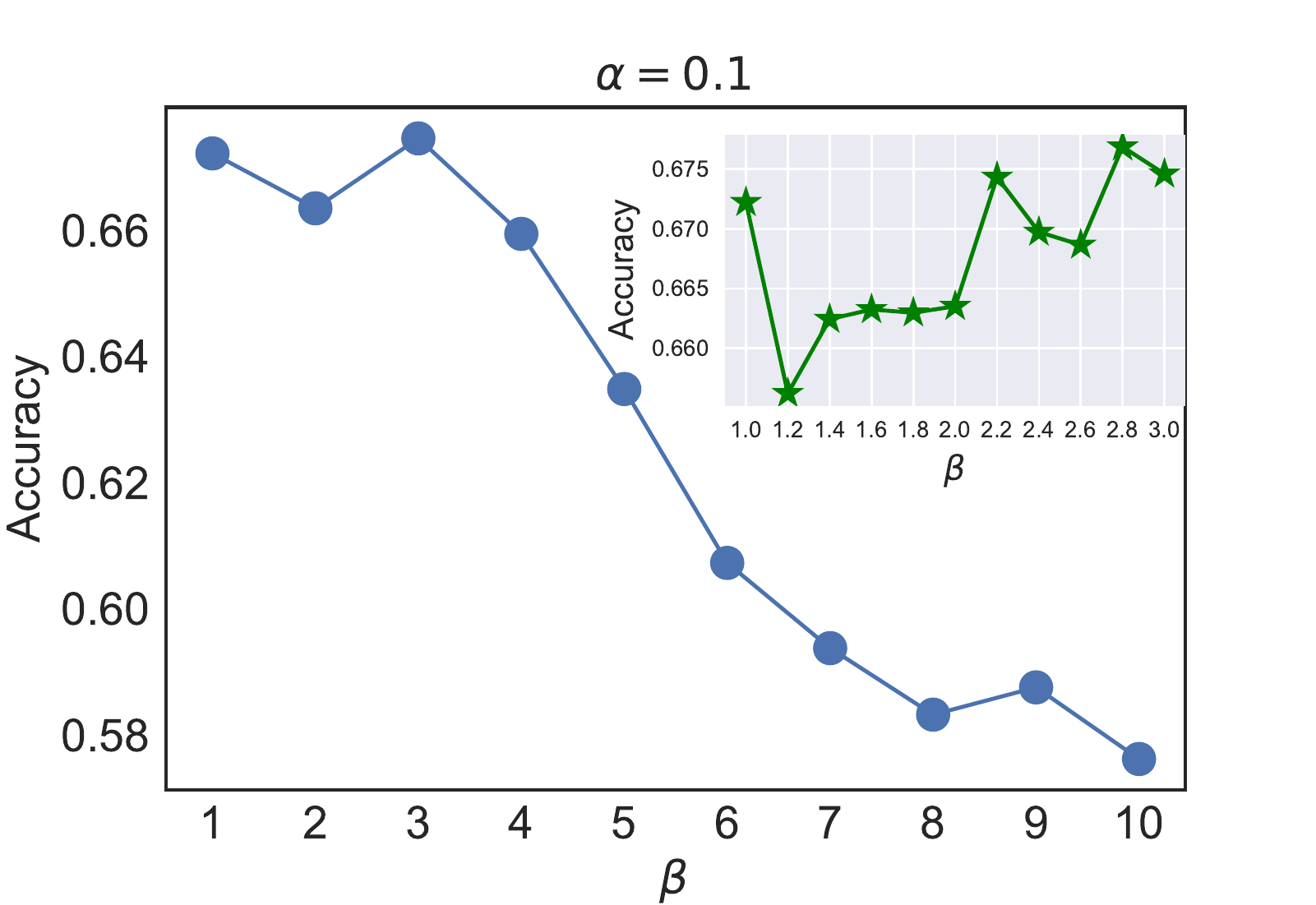}
		    \label{fig:beta_bipolar}
		}%
		\end{subfigure}
		\begin{subfigure}[\emph{HIV}]{
		\includegraphics[width=.75\linewidth]{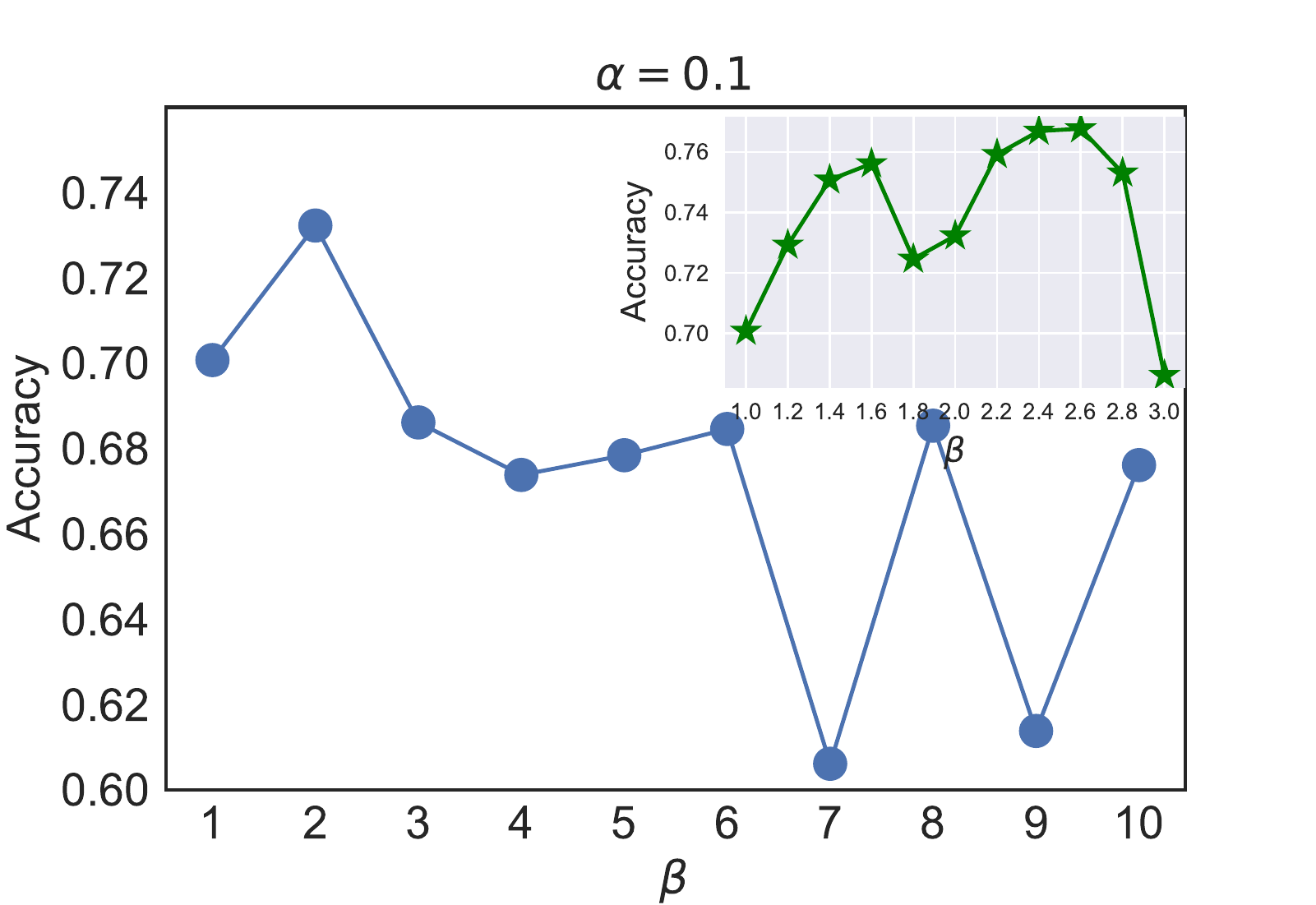}
		\label{fig:beta_hiv}
		}%
		\end{subfigure}  
  \caption{An illustration of the dense grid search for $\beta$. The blue line shows the results of the initial search from the range [1,10] and the green line shows further dense search around the optimal values of the initial search.}\label{fig:beta}
\end{figure}

For the parameters $\alpha$ and $\beta$, we first do a grid search from $\{ {{10^{ - 3}}, \cdots {10^3}}\}$ to find the initial optimal values from this range and then do a further dense search with smaller step size to find the optimal values. Fig.~\ref{fig:para} shows the classification accuracy achieved by SCP-GCN with different values of $\alpha$ and $\beta$ from $\{ {{10^{ - 3}}, \cdots {10^3}}\}$. We can observe that the performance of SCP-GCN is fairly good with most of the parameter values shown in the figures, though some fluctuations may occur. The optimal setting from this range is $\alpha = 0.1, \beta = 1$ for both Bipolar and HIV datasets. We further do a dense search with smaller step size around $\alpha = 0.1, \beta = 1$. Fig.~\ref{fig:beta} shows an illustration of the dense grid search for $\beta$ when $\alpha = 0.1$ on both datasets. The optimal setting found for SCP-GCN on Bipolar is $\alpha = 0.1, \beta = 2.8$ and $\alpha = 0.1, \beta = 2.6$ on HIV.    

\section{Related Works} \label{sec_related}
Our work relates to several branches of studies, which include brain network analysis, Siamese networks, and representation learning on graph. 

Brain network analysis has been an emerging research area, as it yields new insights concerning the understanding of brain function and many neurological disorders \cite{liu2017complex}. 
 Existing works in brain networks mainly focus on discovering brain network from spatio-temporal voxel-level data or mining from brain networks for neurological analysis \cite{ma2016spatio,ma2016multi,bai2017unsupervised,ma2017multi,wang2017structural,bai2018discovering}. For example, in \cite{bai2017unsupervised}, an unsupervised matrix tri-factorization method is developed to simultaneously discover nodes and edges of the underlying brain networks in fMRI data. In \cite{ma2017multi}, the functional network and structural network of each subject are stacked together into a tensor and a tensor factorization based multi-view embedding method is applied to learn a consensus embedding from the two networks for clustering analysis. In \cite{wang2017structural}, a deep model with CNN is proposed to learn non-linear and modular-preserving structures from brain networks for brain disorder diagnosis. In \cite{bai2018discovering}, block models are devised to find structurally and behaviorally feasible block interactions for clustering. Most of these works aim to learn discriminative features from brain networks for the classification or clustering of subjects. Recently, \cite{ktena2018metric,ma2019deep} introduce GCN based similarity learning for brain network analysis. However, they focus on learning a similarity metric on fMRI brain networks, whereas our goal is to jointly learn graph embedding from both DTI and fMRI networks for neurological disorder detection.

Siamese networks were first presented in \cite{bromley1994signature} to investigate image matching problem for signature verification. In image recognition, researchers have been focusing on understanding the mechanism of Siamese architecture and applying this powerful tool \cite{chopra2005learning,koch2015siamese}. Specifically, \cite{chopra2005learning} proposes a CNN-based Siamese framework, which aims to learn effective similarity metric for face verification. Further, to solve one-shot learning problem, \cite{koch2015siamese} proposes a novel generative approach to model unseen classes by leveraging Siamese networks. The Siamese architecture has been successfully utilized in computer vision tasks, but it is rarely explored in the field of graph mining. We aim to take advantage of this architecture to learn the joint embedding of functional and structural brain networks.


Multiple methods have been proposed for learning latent node representations on graph-structured data. DeepWalk \cite{deepwalk} first proposed to learn node embeddings through local information obtained from random walks on graphs. Node2vec \cite{node2vec} extended  DeepWalk  with more  complicated random walk and search methods. In recent years, Graph convolutional network (GCN) has been widely investigated to facilitate graph mining \cite{defferrard2016convolutional,kipf2016semi,liao2019lanczosnet,qu2019gmnn}. In the application of semi-supervised classification for nodes, \cite{kipf2016semi} presents a simple and effective way to learn node embeddings through a re-normalization trick to simplify and speed up computations of previous GCNs. LanczosNet \cite{liao2019lanczosnet} leverages the Lanczos algorithm to construct a low rank approximation of the graph Laplacian, which provides an efficient way to gather multi-scale information for graph convolution.

\section{Conclusions} \label{sec_conclusion}
Heterogeneous sources of brain data provide a valuable opportunity for a more comprehensive  understanding of connectivity and function of the human brain. In this paper, we push forward the analysis of human brain by combining structural connectivity network (e.g., DTI) and functional signals (e.g., fMRI) in a uniform framework, in which the task of accurately classifying brain disease is achieved by incorporating the community-preserving property into graph convolutional networks while learning their structural and functional joint embedding. A pair-wise similarity learning strategy is devised into this unified framework called Siamese Community-Preserving GCN (SCP-GCN). The superiority of our framework is demonstrated by empirical results on two real brain network datasets (i.e., Bipolar and HIV) against state-of-the-art approaches.

\bibliographystyle{IEEEtran}

\end{document}